\documentclass[letterpaper]{article} 
\usepackage[preprint]{aaai2027}  
\usepackage[hyphens]{url}  
\usepackage{graphicx} 
\graphicspath{{figs/}}
\usepackage{natbib}  
\usepackage{caption} 
\usepackage{booktabs}
\usepackage{amsmath}
\usepackage{amssymb}
\usepackage[labelfont=bf]{caption}
\usepackage{floatrow}
\newcommand{\colperp}{\mathrm{col}(W)^{\perp}}
\newcommand{\dmodel}{d_{\mathrm{model}}}
\newcommand{\dff}{d_{\mathrm{ff}}}
\newcommand{\rhoeff}{\rho_{\mathrm{eff}}}
\newtheorem{proposition}{Proposition}

\title{Train What You Deploy: Closing the MLP Reachability Gap\\ in Low-Rank Clone Distillation}
\author{
    Wenhui Chen\textsuperscript{1},
    Zhifeng Li\textsuperscript{1},
    Jie Zhou\textsuperscript{1},
    Navan Preet Singh\textsuperscript{1,2},
    Madalina Ciobanu\textsuperscript{1},\\
    Chenghua Wang\textsuperscript{1},
    Qingqing Mao\textsuperscript{1,2}\corresponding,
    Ritankar Das\textsuperscript{1,2}
}
\affiliations{
    \textsuperscript{1}Incept Labs, Houston, TX \quad
    \textsuperscript{2}Titan Holdings, San Francisco, CA\\
    Correspondence to: Qingqing Mao <qmao@titanholdings.ai>
}

\begin{document}

\thispagestyle{plain}

\maketitle

\begin{abstract}
A compressed student has two shapes that need not agree: the weight it deploys at inference and the weight family its training can reach. We show that a state-of-the-art weight-inheritance distiller, Low-Rank Clone (LRC), deploys a full-width student MLP but ties training to a teacher-induced slice, leaving 62.5--81.4\% of each deployed matrix's independent linear degrees of freedom unreachable---paid for at inference, never trainable. Our principle is one line: train what you deploy. From the identical LRC warm start, we make the training object the entire deployed matrix, with no change in deployed shape, deployed parameter count, or inference FLOPs, via two mergeable realizations (Dense-LRC and CORE-LRC) that both collapse to one deployed weight. This recovers stranded capacity: taking the stronger realization per teacher, $+2.36$/$+2.71$/$+10.45$ Avg9 over matched-budget plain-LRC baselines across three teachers (Llama3.2-3B, Llama3.1-8B, Qwen2.5-3B), with the largest gain on the widest teacher (Qwen), where it reaches the original recipe's $\sim$20B-token accuracy at 10B tokens ($2\times$ token efficiency); there the strictly same-lineage arm still recovers $+6.39$, the fully controlled figure. Controls strongly support attributing the gain to the enlarged reachable set, rather than to added parameters or the recipe. From $\sim$10B distillation tokens plus a short SFT, a half-parameter 1.5B student matches its $\sim$9T-token teacher's 9-task macro-average, within evaluation noise and with a residual MMLU deficit, and a 2.7B student beats Meta's own official compression of Llama3.1-8B at $\sim$900$\times$ fewer compression tokens (a token count under unmatched recipes, not a compute claim). All results are from single-seed runs on the LRC backbone.
\end{abstract}

\section{Introduction}\label{sec:intro}

A well-known discipline in systems is \emph{train/serve consistency}: the pipeline you train should match the one you serve. This paper points out that a state-of-the-art weight-inheritance distiller violates a weight-level version of the same discipline---and that fixing it recovers a large, free gain. A compressed student has two shapes that need not agree: the weight matrix it \emph{deploys} at inference, and the family of matrices its training can actually \emph{reach}. Low-Rank Clone (LRC)~\citep{hao2025lrc} deploys a full-width student MLP but ties its training to a teacher-induced subspace, so most of the deployed matrix's independent degrees of freedom are served yet unreachable to the optimizer. Our fix is the corresponding principle---\emph{train what you deploy}: make the training object the entire deployed matrix.

\begin{figure*}[t]
\centering
\includegraphics[width=1\textwidth]{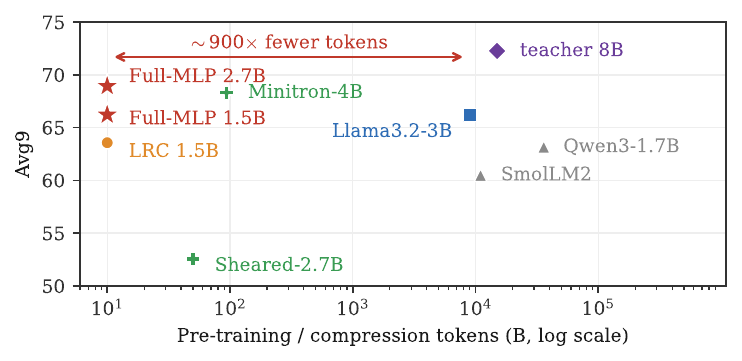}
\caption{Token efficiency of training LRC's full deployed MLP matrix. Avg9 ($0$-shot) versus training tokens (log scale). From $\sim$10B \emph{distillation} tokens (plus a short SFT), our students (stars; stronger realization per setting: teacher-spectral $66.21$ at $1.5$B, canonical dense $68.93$ at $2.7$B) reach their $\sim$9T-token teacher (Llama3.2-3B) and Meta's official same-lineage compressions; the $2.7$B student clears the official Llama3.2-3B at $\sim$900$\times$ fewer \emph{compression} tokens (a token count under unmatched recipes; both pay the teacher's own pre-training). From-scratch SLMs (Qwen3-1.7B, SmolLM2) are size context only; external points are published values under the same harness (\S\ref{sec:limitations}).}
\label{fig:frontier}
\end{figure*}

From the identical LRC warm start, at no change in deployed shape, this lets a half-parameter $1.5$B student match its $\sim$9T-token teacher (Llama3.2-3B) on the 9-task macro-average, a $2.7$B student beat Meta's official same-lineage compression at $\sim$900$\times$ fewer compression tokens (a compression-stage token count under unmatched recipes), and---on the widest teacher, where the constraint strands the most---a $1.7$B student gain $+10.45$ Avg9 over the matched $10$B-token baseline, reaching the original recipe's $\sim$20B-token accuracy at half the tokens; all at \emph{zero} added inference cost (Figure~\ref{fig:frontier}).

\paragraph{An audit any compressor can fail.}
For a deployed weight, define its \emph{training utilization} $u$ as the ratio of its training-reachable dimension to its deployed dimension. A gap ($u<1$) is only justified if it buys something: LoRA-style adapters accept $u<1$ to \emph{protect} pretrained content; true structured pruning accepts it because the deleted dimensions are also removed from deployment. LRC's MLP gap buys neither. It compresses only the hidden dimension and \emph{inherits} the teacher's MLP intermediate width $\dff$: it deploys a $\dff\times\dmodel^{(S)}$ matrix but writes only into a $\dmodel^{(T)}$-dimensional teacher column slice, so $u=\dmodel^{(T)}/\dff=1/\rho$. Because a modern MLP is wide---our three teachers span $\rho=2.67$ to $5.375$---this leaves $1-1/\rho=62.5$--$81.4\%$ of the deployed matrix's independent linear degrees of freedom unreachable to training, at full deployment cost (what is stranded is independent \emph{directions}, not fixed entries; formalized in \S\ref{sec:observations}). Recent spectral work~\citep{jha2025spectral} argues much of that nominal width is never effectively used; but ``rarely used at initialization'' and ``useless once trained'' are different claims, and only an intervention can tell them apart.

\begin{figure*}[t]
\centering
\includegraphics[width=\textwidth]{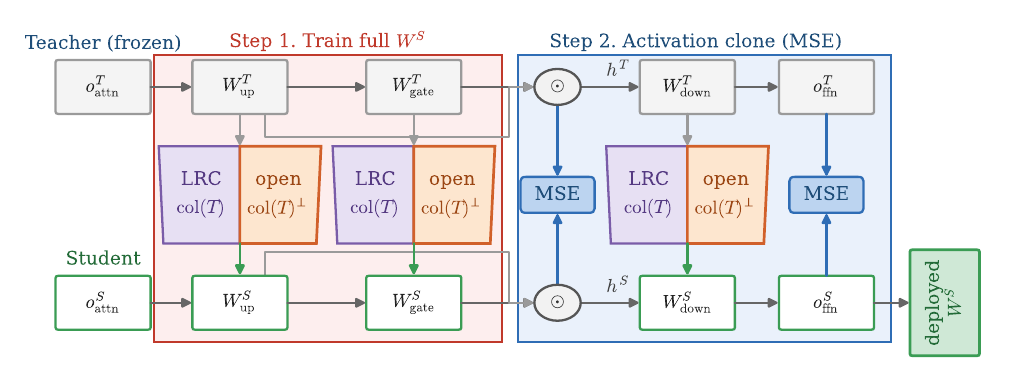}
\caption{The Full-MLP LRC framework. Full-MLP LRC trains the entire deployed MLP weight $W^{S}$ from the LRC warm start (Step 1): the teacher column space $\mathrm{col}(T)$ (purple), which LRC already trains (the warm start), plus the structural complement $\mathrm{col}(T)^{\perp}$ (orange, $\dim{=}\dff-\dmodel^{(T)}$) that unfreezing now opens. The student clones the teacher's intermediate activations and FFN outputs by per-layer MSE (Step 2); $W^{S}$ keeps its original shape, so the deployed model adds no parameters or FLOPs over plain LRC. CORE-LRC opens the orange path as a zero-initialized, mergeable teacher-basis path (Eq.~\ref{eq:ocd}), functionally identical to the dense continuation. Training: teacher-guided distillation (temperature-scaled logit KL, per-layer hidden/attention MSE) with a next-token CE regularizer.}
\label{fig:framework}
\end{figure*}

\paragraph{Train what you deploy.}
Acting on the principle is the simplest possible intervention: train the full deployed $\dff\times\dmodel^{(S)}$ matrix from the LRC warm start, initialized so the student starts \emph{exactly} at the LRC model and merges back to the identical deployed shape (Figure~\ref{fig:framework}). We realize it in two \emph{mergeable} ways: a plain dense weight trained in full from the merged LRC warm start (\textbf{Dense-LRC}), and a teacher-spectral reparameterization of the same full matrix (\textbf{CORE-LRC}, \S\ref{sec:method})---both merging to the identical deployed weight at zero added inference cost. Controls (\S\ref{sec:design}) place the cause on the reachable set itself: a canonical-dense arm, a random ambient completion, and the teacher-basis reparameterization all recover the gain on the narrow Llama teachers, and the merged weight keeps the standard deployed shape and deployed parameter count. A fully equal-parameter arm whose added coordinates are confined to the teacher slice---expanding no reachable set while adding the \emph{same} trainable coordinates and optimizer state---recovers \emph{none} of the gain (within noise of plain LRC); and the slice$\to$full gain persists under a stripped \emph{generic} recipe (logit-KL only, random initialization), so it is recipe-independent, not an artifact of the LRC objective. On the widest, most ill-conditioned teacher the two realizations separate: the teacher-spectral basis realizes far more of the \emph{same} expansion at a fixed budget ($+10.45$ vs.\ $+6.39$ Avg9; \S\ref{sec:experiments}), making CORE-LRC the realization of choice where compression is hardest (basis-conditioning analysis in the supplementary material).

\paragraph{Contributions.}
\begin{itemize}
\item \textbf{A deployment--training reachable-set gap in LRC, made auditable.} We frame a compressed weight by a training-utilization ratio $u$ and show LRC's MLP sits at $u=1/\rho$: the deployed family $\{TZ^{\top}\}$ is a strict subset of $\mathbb{R}^{\dff\times r}$, stranding $62.5$--$81.4\%$ of each deployed matrix's independent degrees of freedom at full inference cost and no deployment saving---a gap that, unlike LoRA's or a true prune's, buys nothing (\S\ref{sec:observations}).
\item \textbf{Train what you deploy, with controlled attribution.} Training the full deployed matrix from the identical warm start recovers the stranded capacity at zero added inference cost: $+2.36$/$+2.71$/$+10.45$ Avg9 over matched-budget plain-LRC baselines across three teachers (stronger realization per teacher, Table~\ref{tab:unified}; the strictly same-lineage dense arm alone gives $+2.23$/$+2.71$/$+6.39$), reaching $2\times$ token efficiency on the widest teacher, with a half-size $1.5$B student matching its teacher's macro-average. Controls place the cause on the enlarged set: full-matrix parameterizations recover it, the \emph{deployed} shape and parameter count are unchanged (training-time trainables rise from $rH$ to $r\dff$ per projection---exactly matched by the equal-parameter control), a long-budget sweep argues against mere faster convergence, an equal-parameter slice-confined arm recovers \emph{none} of it, the gain persists under a stripped generic recipe, and occupancy/knockout show the opened directions are used (\S\ref{sec:design}).
\end{itemize}

\begin{table}[t]
\centering
{\small
\setlength{\tabcolsep}{2pt}
\begin{tabular}{lccc}
\toprule
Setting ($\rho^{(T)}$) & Plain LRC & Full-MLP LRC & $\Delta$ \\
\midrule
Llama3.2-3B$\to$1.5B ($2.67$) & $63.04$ & $65.40$ & $+2.36$ \\
Llama3.1-8B$\to$2.7B ($3.50$) & $65.55$ & $68.26$ & $+2.71$ \\
Qwen2.5-3B$\to$1.7B ($5.375$) & $52.99$ & $63.44$ & $+10.45$ \\
\bottomrule
\end{tabular}}
\caption{Training the deployed matrix vs.\ a matched plain-LRC baseline ($10$B PT, Avg9, $0$-shot; Avg9 is the mean over $9$ tasks including MMLU, excluding MathQA). Full-MLP LRC has two mergeable realizations---canonical Dense-LRC and teacher-spectral CORE-LRC---that merge to the identical deployed weight at zero added inference cost and \emph{tie} on Llama (basis-independence); each row reports the stronger per teacher (teacher-spectral on Llama3.2-3B and Qwen, dense on Llama3.1-8B). On the widest teacher (Qwen) the teacher-spectral realization reaches $63.44$, matching the original $\sim$20B-token baseline at half the tokens; it is a separate training lineage from the matched pair, whose same-lineage dense arm recovers $+6.39$ ($59.38$)---the fully controlled attribution evidence (\S\ref{sec:experiments}).}
\label{tab:unified}
\end{table}

\section{The Inherited Width and Its Stranded Complement}\label{sec:observations}

LRC compresses only the hidden dimension $\dmodel$ and \emph{inherits} the teacher's full intermediate width $\dff$, yet trains only a $\dmodel^{(T)}$-dimensional teacher slice of each MLP projection. Our three teachers span $\rho=\dff/\dmodel$ of $2.67$, $3.50$, and $5.375$, so the inherited-but-untrained part is large and teacher-dependent. For a gate/up projection $W\in\mathbb{R}^{\dff\times\dmodel}$ of full column rank, the training-inaccessible orthogonal complement has dimension $d_{\perp}=\dff-\dmodel$, a fraction
\begin{equation}\label{eq:fperp}
f_{\perp} = \frac{d_{\perp}}{\dff} = 1 - \frac{1}{\rho}
\end{equation}
of the intermediate space: $62.5\%$ at $\rho=2.67$, $71.4\%$ at $3.50$, and $81.4\%$ at $5.375$. This bounds excluded \emph{matrix degrees of freedom} of one non-square projection, not unused neurons (the SwiGLU product mixes gate and up outputs). Whether these directions are useless or unused-but-usable is settled only by opening them and measuring what is recovered---the intervention this paper makes.

\begin{proposition}[Reachability gap]\label{prop:gap}
Let $\mathcal{H}_{\mathrm{deploy}}{=}\mathbb{R}^{\dff\times r}$
be one student projection's deployment hypothesis
space ($r{=}\dmodel^{(S)}$) and
$\mathcal{R}_{\mathrm{LRC}}{=}\{TZ^{\top}\}\subsetneq\mathcal{H}_{\mathrm{deploy}}$
the reachable set of the LRC parameterization ($T$
frozen). The \emph{reachability gap}
$\mathcal{G}{=}\mathcal{H}_{\mathrm{deploy}}\setminus\mathcal{R}_{\mathrm{LRC}}$
is deterministic: fixed by the parameterization, for
any initialization, objective, or optimizer.
\end{proposition}

Reachability concerns \emph{optimization accessibility}, not representational capacity: the deployed shape carries any $W\in\mathcal{H}_{\mathrm{deploy}}$ at identical inference cost, yet training only ever produces a member of $\mathcal{R}$. Membership in $\mathcal{R}$ is optimizer-independent; how much of $\mathcal{R}$ a \emph{finite-budget} optimizer attains also depends on its coordinates (AdamW is not rotation-invariant;~\citealp{zhang2025adamrotation})---the first level carries the structural claim (\S\ref{sec:design}), the second surfaces only on the widest teacher (\S\ref{subsec:qwen}).

\section{Related Work}\label{sec:related}

\paragraph{Compression-distillation and pruning.}
Knowledge distillation~\citep{hinton2015distilling} underlies DistilBERT~\citep{sanh2019distilbert}, TinyBERT~\citep{jiao2020tinybert}, MiniLM~\citep{wang2020minilm}, and sequence-level objectives for generative LLMs~\citep{gu2024minillm,agarwal2024gkd}. Low-Rank Clone~\citep{hao2025lrc}, the backbone we build on, jointly soft-prunes teacher weights by low-rank projection and clones teacher activations, matching trillion-token-trained models with about $20$B tokens. A complementary line \emph{removes} capacity: structured pruning with distillation-based retraining cuts depth and width, including the MLP intermediate dimension, as in Minitron~\citep{muralidharan2024minitron} and Sheared LLaMA~\citep{xia2024sheared}. This contrast is our starting point: where width pruning \emph{deletes} the intermediate dimension, LRC \emph{inherits} the teacher's $\dff$ against a smaller $\dmodel$; we treat the resulting structural complement as a reserve to fill rather than waste.

\paragraph{Concurrent diagnoses, and the growth dual.}
RED~\citep{he2026red} diagnoses a different failure in the same family of projection distillation: effective-rank collapse of hidden representations, repaired by an activation-aware channel-selection \emph{initialization}, after which training still proceeds \emph{within} the projected family $\{TZ^{\top}\}$. We instead identify a \emph{deterministic parameterization gap} (the $\mathcal{G}$ of \S\ref{sec:observations}) and remove the family constraint itself. The two axes are empirically separable in our data: an activation-aware channel-passthrough initialization of that flavor is worth $+0.68$ Avg9 on the Qwen target, whereas opening the reachable set at the same budget recovers $+6.39$ to $+10.45$ (\S\ref{subsec:qwen}); the representation-rank statistic dissociates from the gain (\S\ref{subsec:engagement}). Spectral scaling analyses~\citep{jha2025spectral,jha2026capacity} report that FFN width is under-utilized and usable capacity is optimizer-dependent; these are \emph{observational}, and we supply the constructive counterpart by intervention. From the opposite direction, function-preserving growth~\citep{samragh2024hypercloning} expands a small model and relies on full training to activate initially unused directions; our setting is the compression dual.

\paragraph{Subspace and null-space adapters.}
Mechanically, the teacher-spectral realization (Eq.~\ref{eq:ocd}) is a constrained SVD-defined subspace update, as in PEFT methods~\citep{meng2024pissa,wang2024milora,liu2024dora,tang2025loranull,xiong2026oplora}; we claim no novelty in the mechanism, and the gain does not depend on it: several full-matrix parameterizations recover it (\S\ref{subsec:dense_control}). The setting differs: those methods \emph{avoid disturbing} pretrained knowledge during fine-tuning, whereas we \emph{add} usable capacity during distillation pre-training, targeting the structural left null space of a non-square \emph{inherited} weight and merging back at inference.

\paragraph{Is low-rank enough for distillation?}
A concurrent line argues low-rank distillation is sufficient, even optimal~\citep{soarez2026demystifying,kalyoncuoglu2025highdim,shen2026opdgeometry}. No conflict: those concern the \emph{rank of the trainable update} within a fixed parameterization; our gap is the \emph{inherited} $\dff$ that LRC retains but never opens, on which confining the update to the teacher subspace is demonstrably not sufficient. We make no claim that low-rank is inferior in general.

\section{Train What You Deploy}\label{sec:method}

A non-square MLP weight splits its output space into the teacher-occupied range $\mathrm{col}(W)$ and a structural complement $\colperp$ that LRC's projection never opens (for a gate/up weight this is the \emph{left} orthogonal complement $\mathrm{null}(W^{\top})$; for the down projection, the right one). \emph{Training what we deploy} opens it: we train the \emph{entire} deployed $\dff\times r$ MLP matrix from the plain-LRC warm start, at no change in deployed shape, parameter count, or FLOPs.

\subsection{Dense-LRC: The Construction}\label{subsec:construct}
Let $W\in\mathbb{R}^{\dff\times H}$ be a teacher gate or up projection with $\dff>H$, where $H{=}\dmodel^{(T)}$ and $r{=}\dmodel^{(S)}$ is the (smaller) student width. Plain LRC parameterizes its compressed projection as $W_{\mathrm{LRC}}=T\,Z_{\mathrm{col}}^{\top}$, with $T$ an orthonormal basis of $\mathrm{col}(W)$ and only the $r\times H$ coordinates $Z_{\mathrm{col}}$ trained, confining its column space to $\mathrm{col}(W)$ (the reachable set $\mathcal{R}_{\mathrm{LRC}}$ of \S\ref{sec:observations}). \textbf{Dense-LRC} instead parameterizes the projection as a plain dense $\dff\times r$ weight, \emph{initialized at the merged plain-LRC weight} (the student is exactly the plain-LRC model at step zero), and trains every entry; its reachable column space is the full $\mathbb{R}^{\dff}$. At inference the trained weight is a single $\dff\times r$ matrix of the plain-LRC shape---a structural reparameterization in the spirit of inference-time branch merging~\citep{ding2021repvgg}, with no added parameters or FLOPs. An equivalent \emph{teacher-spectral} realization, \textbf{CORE-LRC}, opens the same complement in the teacher's own SVD basis: with $U_\perp\in\mathbb{R}^{\dff\times(\dff-H)}$ an orthonormal basis of $\colperp$,
\begin{equation}\label{eq:ocd}
W_{\mathrm{student}} = T\,Z_{\mathrm{col}}^{\top} + U_\perp\,Z_\perp^{\top},
\end{equation}
with $Z_\perp\in\mathbb{R}^{r\times(\dff-H)}$ zero-initialized, so training also starts exactly at the plain-LRC model and merges to the same single $\dff\times r$ weight (mergeability proposition in the supplementary material). Because $[\,T\ U_\perp\,]$ is a complete orthonormal basis of $\mathbb{R}^{\dff}$, CORE-LRC trains the same standard weight in a fixed teacher-aligned coordinate system---functionally equivalent to Dense-LRC, differing only in the coordinates AdamW optimizes in. On the matched Llama settings the two tie (\S\ref{subsec:dense_control}); on the widest, most ill-conditioned teacher the teacher-spectral coordinates are decisively better ($+10.45$ vs.\ $+6.39$; \S\ref{sec:experiments}). Training it \emph{removes} LRC's $\mathrm{col}(T)$ constraint rather than overwriting the cloned weight; the performance recovered then measures how much of the excluded capacity is usable, and on the Llama targets it is basis-independent (canonical-dense, random-completion, and teacher-basis endpoints within $\sim$0.64 Avg9; \S\ref{subsec:dense_control}). Opening the complement is \emph{not} a zero-forgetting guarantee: single-layer orthogonality does not control the end-to-end Jacobian, and MMLU does drop relative to the teacher.

\subsection{Instantiation and What We Claim}\label{subsec:instantiation}
For Llama3.2-3B$\to$1.5B each MLP weight is $8192\times1536$; Dense-LRC trains this full matrix from the merged warm start. We open the gate, up, and down projections \emph{only}; all other LRC system switches are disabled in every reported run, so the measured gain is attributable to the MLP alone (the supplementary material). Training is distillation against the frozen teacher under a composite objective: temperature-scaled KL on logits (dominant), per-layer hidden-state and attention MSE alignment ($\lambda_{\mathrm{aux}}{=}0.2$), and a next-token cross-entropy regularizer. Dense-LRC \emph{is} warm-started dense continuation of the deployed MLP weight, so our contribution is not that it beats dense training; it is a \emph{diagnosis and repair of a reachable-set gap in LRC}: (i) we formalize the gap; (ii) we show by intervention that opening it stably helps across three compressions; (iii) the fix is free at inference and never perturbs the warm start; and (iv) the teacher-aligned coordinate split makes ``how much optimization leaves the inherited subspace'' measurable (\S\ref{subsec:engagement}).

\section{Experiments}\label{sec:experiments}

\subsection{Setup}\label{subsec:setup}
We compress three teachers with LRC, then train the full deployed MLP matrix from the merged plain-LRC warm start during pre-training (PT). \emph{Realizations and pairing discipline:} ``Full-MLP LRC'' names the family; its two realizations merge to identical deployed weights and tie under matched conditions on the Llama teachers (\S\ref{subsec:dense_control}). Headline deltas take the stronger of the two realizations minus the matched $10$B plain-LRC baseline ($+2.36$/$+2.71$/$+10.45$: teacher-spectral on Llama3.2-3B and Qwen, dense on Llama3.1-8B). The canonical dense arm alone gives $+2.23$/$+2.71$/$+6.39$; on Qwen the dense delta is the strictly same-lineage matched one (the CORE-Qwen arm is a separate lineage), so the controlled attribution never rests on a cross-lineage number. Both are reported and labeled wherever they appear. Evaluation uses nine standard tasks plus MathQA via the LM Evaluation Harness~\citep{gao2024lmeval}, all \emph{standard $0$-shot} (MMLU included), matching the LRC protocol; all models are official Instruct checkpoints under one harness. Matched deltas pair both arms at one stage (PT; also SFT for the 3B setting, $+2.38$ Dense); teacher and reference comparisons use the SFT student, matching the Instruct references. PT data follows the original LRC recipe ($10$B FineWeb-Edu~\citep{penedo2024fineweb} $+$ $0.35$B OpenHermes tokens); an optional short SFT adds $0.62$B tokens. Our pipeline is near-deterministic (deterministic data-aware SVD initialization; data order fixed by a shuffle seed), and our plain-LRC baseline reproduces LRC's published numbers. We report \emph{single-seed} training runs (the field norm at this scale); run-to-run variation is bounded by the lm-eval Avg9 standard error (at most $\approx$$0.55$ across our
settings, cleared $\sim$4--19$\times$ by the gains) and directional consistency across three teachers---though neither replaces a multi-seed sweep (\S\ref{sec:limitations}). Code, configurations, and run logs are available from the corresponding author on request.

\subsection{Gains on Three Teachers}\label{subsec:main}
\begin{figure}[t]
\centering
\includegraphics[width=1\columnwidth]{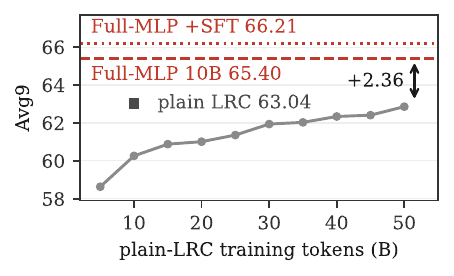}
\caption{Plain LRC saturates below Full-MLP LRC. A no-complement LRC run swept to $50$B tokens (grey; a larger, more diverse corpus) plateaus at $62.86$ Avg9, and the matched plain-LRC point ($63.04$ at $10$B) does not improve under continued training---both $\sim$2.4--2.5 below Full-MLP LRC at $10$B ($65.40$ PT; $66.21$ with SFT $\approx$ teacher $66.18$). A genuine capacity gain, not faster convergence.}
\label{fig:ceiling}
\end{figure}

\paragraph{Llama3.2-3B$\to$1.5B (primary).}
The cleanest comparison fixes the paradigm and toggles only whether the MLP is unfrozen (Table~\ref{tab:llama3}). Our no-complement LRC baseline reproduces the published LRC numbers closely ($63.04$/$63.57$ PT/SFT vs.\ their $62.48$/$63.48$), so the gain is measured against a faithful, same-data reproduction at the original recipe's budget. Full-MLP LRC raises the 9-task average by $+2.36$ to $+2.64$ and MMLU by $+4.02$ (teacher-spectral realization; the matched Dense-LRC arm lands within noise at $+2.23$, Table~\ref{tab:ctrl}); gains concentrate on knowledge/reasoning tasks (CSQA $+6.30$, MMLU $+4.02$). A long-budget \emph{no-complement} control shows this is not faster convergence (Figure~\ref{fig:ceiling}): continuing the matched plain-LRC student past $10$B does not improve it ($63.04\!\to\!63.04$), and an \emph{independent} plain-LRC run trained end-to-end under its own schedule to $50$B tokens on a larger, more diverse corpus plateaus at $62.86$---both $\sim$2.4--2.5 Avg9 below Full-MLP LRC's $65.40$ at $10$B, and at no budget or corpus we tried did plain LRC reach it: a genuine capacity gain. After a short SFT the $1.5$B student (teacher-spectral realization) reaches $66.21$ Avg9, statistically indistinguishable from the $3$B teacher's $66.18$ on this macro-average ($+0.03$, far inside this setting's $\approx$0.4 Avg9 eval SE)---a match on the aggregate, not task by task. The match is not specific to that realization: a \emph{Dense-LRC} student (independent seed-1234 control lineage) reaches $65.95$ after the same short SFT, likewise within evaluation noise ($\Delta{=}{-}0.23$; Table~\ref{tab:llama3}). The profile is \emph{differentiated}, not a uniform tax: the student exceeds its teacher on most commonsense/QA tasks (CSQA $+7.21$, ARC-E $+2.49$, BoolQ $+1.62$; WinoGrande $-4.02$ the exception) while the deficit concentrates in broad-knowledge MMLU ($-5.70$), which recovers only as $\dmodel^{(S)}$ grows (supplementary material).

\begin{table*}[t]
\centering
{\small
\setlength{\tabcolsep}{3.4pt}
\begin{tabular}{lccccccccccc}
\toprule
Model & ARC-E & ARC-C & LogiQA & CSQA & PIQA & WinoG & BoolQ & SciQ & MMLU & \textbf{Avg9} & MathQA \\
\midrule
Teacher Llama3.2-3B & 73.86 & 46.25 & 29.65 & 67.97 & 75.56 & 68.03 & 78.53 & 95.30 & 60.56 & \textbf{66.18} & 34.84 \\
\midrule
LRC-1.5B PT (orig.\ LRC)  & 73.40 & 42.15 & 31.03 & 64.46 & 71.60 & 61.88 & 73.27 & 94.40 & 50.09 & 62.48 & -- \\
LRC-1.5B PT$+$SFT (orig.\ LRC) & 74.75 & 44.97 & 30.72 & 65.77 & 73.07 & 62.25 & 75.78 & 94.60 & 49.42 & 63.48 & -- \\
LRC-1.5B 10B PT (ours)    & 74.12 & 42.66 & 31.34 & 65.36 & 71.93 & 62.43 & 73.73 & 95.10 & 50.65 & 63.04 & 28.84 \\
LRC-1.5B 10B PT$+$SFT (ours) & 74.87 & 43.69 & 30.41 & 67.32 & 73.01 & 61.09 & 75.69 & 95.20 & 50.82 & 63.57 & 29.75 \\
\midrule
Full-MLP LRC 1.5B 10B PT        & 76.43 & 46.08 & 30.41 & 71.66 & 74.10 & 64.56 & 75.57 & 95.10 & 54.67 & 65.40 & 30.42 \\
Full-MLP LRC 1.5B 10B PT$+$SFT  & 76.35 & 45.31 & 30.26 & 75.18 & 74.65 & 64.01 & 80.15 & 95.10 & 54.86 & \textbf{66.21} & 31.86 \\
Full-MLP LRC 1.5B 16.5B PT$+$SFT& 75.72 & 44.11 & 30.88 & 75.10 & 74.32 & 63.69 & 82.11 & 95.20 & 55.38 & \textbf{66.28} & 32.46 \\
\midrule
Dense-LRC 1.5B 10B PT$+$SFT\textsuperscript{\S} & 75.59 & 44.37 & 29.65 & 73.87 & 73.78 & 65.11 & 81.47 & 94.90 & 54.81 & 65.95 & 31.49 \\
\midrule
$\Delta$ (Full-MLP$-$LRC, 10B PT)   & +2.31 & +3.42 & $-$0.93 & +6.30 & +2.17 & +2.13 & +1.84 & 0.00 & +4.02 & \textbf{+2.36} & +1.58 \\
$\Delta$ (Full-MLP$-$LRC, 10B PT$+$SFT) & +1.48 & +1.62 & $-$0.15 & +7.86 & +1.64 & +2.92 & +4.46 & $-$0.10 & +4.04 & \textbf{+2.64} & +2.11 \\
\bottomrule
\end{tabular}}
\caption{Llama3.2-3B$\to$1.5B: Full-MLP LRC vs.\ a matched no-complement LRC baseline, the original LRC paper, and the teacher (re-evaluated under our harness; all $0$-shot; Avg9 excludes MathQA). Full-MLP rows: teacher-spectral realization; the matched Dense-LRC arm reaches $65.27$ PT ($+2.23$, Table~\ref{tab:ctrl}). \textsuperscript{\S}Dense-LRC$+$SFT from the independent seed-1234 lineage (\S\ref{subsec:dense_control}): $65.95$ vs.\ teacher $66.18$ ($\Delta{=}{-}0.23$, inside eval SE)---the teacher match is carried by the canonical realization too. ``$n$B'' is always the \emph{PT} budget; ``$+$SFT'' is the short $0.62$B-token 8-dataset instruction stage (supp.).}
\label{tab:llama3}
\end{table*}

\paragraph{Llama3.1-8B$\to$2.7B.}
The same matched comparison on a larger teacher confirms the gain is not specific to the 3B model: training the deployed matrix adds $+2.71$ Avg9 and $+4.55$ MMLU over the LRC baseline at 10B PT ($65.55\!\to\!68.26$; Dense-LRC realization); after a short SFT the $2.7$B student reaches $68.93$ (per-task in Table~\ref{tab:llama8}). A sharper reading compares this student against \emph{Llama3.2-3B}, Meta's own prune--distill compression of this very teacher trained on up to $\sim$9T tokens~\citep{grattafiori2024llama3}. Full-MLP LRC compresses the same Llama3.1-8B with only $10$B PT tokens---roughly $900\times$ fewer---into a slightly smaller student ($\approx$2.7B vs.\ $3.21$B). Since the reference is instruction-tuned, the like-for-like comparison is our SFT student: $68.93$ vs.\ $66.18$ ($+2.75$; the PT student already leads at $68.26$), trailing on MMLU ($58.54$ vs.\ $60.56$) and WinoGrande while exceeding on the other seven tasks. Because the reference shares the teacher but not the recipe, data, or post-training, this is a same-lineage token-count comparison---not a controlled recipe comparison or a compute claim (only the compression-stage budget is counted, a convention both sides share; \S\ref{sec:limitations}). In external context (Table~\ref{tab:external}), the $2.7$B Dense-LRC student ($68.93$) also exceeds the same-size Sheared-LLaMA-2.7B ($52.55$) and the larger Minitron-4B ($68.31$) at $\sim$9$\times$ fewer tokens (published numbers, context only).

\paragraph{The hard wide-MLP target: Qwen2.5-3B$\to$1.7B.}\label{subsec:qwen}
The Qwen2.5-3B teacher ($\rho^{(T)}{=}5.375$) is a structurally harder compression target, visible before any full-matrix training: the original LRC recipe trains this student with $\sim$20B tokens, twice the Llama budget. Under a matched $10$B budget our plain-LRC baseline reaches only Avg9 $52.99$ (MMLU $36.42$), severely under-converged---and this widest MLP is where training what you deploy pays off most. The teacher-spectral CORE-LRC realization reaches Avg9 $\mathbf{63.44}$ at $10$B PT ($63.82$ after SFT)---$\mathbf{+10.45}$ over the matched baseline, \emph{matching LRC's own $\sim$20B-token baseline ($63.43$) at half the tokens} ($2\times$ token efficiency), with the gain landing where plain LRC most under-fits (CSQA $+31.12$, MMLU $+17.03$). The canonical Dense-LRC realization, trained in the \emph{strictly same lineage} as the baseline (same seed, initialization, and data order), independently recovers $\mathbf{+6.39}$ ($59.38$; CSQA $+21.62$, MMLU $+13.71$): the fully controlled comparison establishes the reachable-set gain on its own. As a control against an initialization-side account~\citep{he2026red}, an activation-aware channel-passthrough initialization is worth only $+0.68$ Avg9 here---an order of magnitude smaller than opening the reachable set, which it neither explains nor bounds. On this widest, most ill-conditioned target the teacher-spectral basis is the markedly better optimizer ($\approx$4 above Dense; the two tie on Llama). The CORE-Qwen arm is a separate lineage from the seed-matched pair, so we read that margin qualitatively; the anchors needing no such reading are the matched $+6.39$ and CORE reaching the $20$B recipe's accuracy at $10$B (the supplementary material).

\section{Isolating the Reachable Set}\label{sec:design}

Three alternatives must be excluded before attributing the gain to the enlarged reachable set---a privileged basis, a pure added-parameter effect, an LRC-recipe artifact. Each matched control toggles one ingredient of \S\ref{sec:observations}'s definition: basis controls change \emph{coordinates}, not $\mathcal{R}$; the equal-parameter arm adds trainable coordinates, not $\mathcal{R}$; the cross-recipe $2\times2$ changes the \emph{objective}, not the $\mathcal{R}$ contrast.

\subsection{Basis Controls: Capacity, Not Coordinates}\label{subsec:dense_control}
We re-train the primary Llama3.2-3B$\to$1.5B student with each MLP projection as a plain dense weight (canonical basis), from the \emph{same} merged warm start under the identical recipe (Table~\ref{tab:ctrl}). The dense control reaches $65.27$ Avg9 versus the teacher-spectral $65.40$ ($\Delta{=}{-}0.13$, within run-to-run variation; under an independent data-shuffle seed the two arms again land within noise, $\Delta\!\approx\!0.15$): the basis is not privileged. A \emph{random ambient completion} arm (random full-rank completion, not orthogonalized against $\mathrm{col}(T)$; reachable set again all of $\mathbb{R}^{\dff\times r}$) reaches $64.76$ under the same recipe, seed, and data order. All three parameterizations of the same unlocked capacity recover the bulk of the gain (spread $0.64$ Avg9, the non-orthogonalized random completion lowest): the capacity, not the coordinates, carries the gain. The trained weights corroborate this: the dense control ends with $7.1\%$/$9.4\%$ (gate/up) of its MLP weight energy in $\colperp$, matching the teacher-spectral student's $6.2\%$/$8.3\%$---the objective, not the parameterization, determines where the solution lives.

\begin{table}[t]
\centering
{\small
\setlength{\tabcolsep}{1mm}
\begin{tabular}{lccc}
\toprule
 & LRC & CORE & Dense \\
\midrule
Train./proj. & $rH$ & $r\dff$ & $r\dff$ \\
Total MLP train. & $0.40$B & $1.06$B & $1.06$B \\
Reachable set & $\mathrm{col}(T)$ & $\mathbb{R}^{\dff}$ & $\mathbb{R}^{\dff}$ \\
Warm start & LRC init & $=$LRC init & $=$LRC init \\
Avg9 @ $10$B PT & $63.04$ & $65.40$ & $65.27$ \\
\bottomrule
\end{tabular}}
\caption{Capacity accounting for the basis controls (Llama3.2-3B$\to$1.5B; $H{=}3072$, $r{=}1536$, $\dff{=}8192$, $28$ layers). CORE and the dense control train the same standard weight from the same warm start and deploy identically, differing only in coordinates.}
\label{tab:ctrl}
\end{table}

\subsection{The Equal-Parameter Control: Capacity, Not Parameter Count}\label{subsec:redundant}
The decisive control adds the \emph{same} $r\dff$ trainable coordinates per projection as the full-matrix arm but confines them to the teacher slice: $W_{\mathrm{red}}=T\,(Z_{\mathrm{col}}^{\top}+A\,Z_{\mathrm{red}}^{\top})$ with $Z_{\mathrm{col}}\in\mathbb{R}^{r\times H}$, $Z_{\mathrm{red}}\in\mathbb{R}^{r\times(\dff-H)}$ both trainable and $A\in\mathbb{R}^{H\times(\dff-H)}$ a \emph{fixed}, unit-column random matrix. The count is exactly the full-matrix arm's, with equal AdamW state, while $\mathrm{col}(W_{\mathrm{red}})\subseteq\mathrm{col}(T)$ by construction. Trained from the same warm start under the identical recipe (matched seed-1234 lineage), it reaches Avg9 $64.02$---within noise of plain LRC ($64.04$ in this lineage) and $\sim$1.1 points below the full-matrix arm in the same lineage ($65.16$). The same coordinate and optimizer-state budget, absent reachable-set expansion, recovers essentially none of the gain, strongly supporting reachable-set expansion as the cause (``strongly supports,'' not ``decisively isolates''; caveats in \S\ref{sec:limitations}).

\subsection{Recipe-Independence}\label{subsec:crossrecipe}
The controls above hold the LRC recipe fixed (activation/attention alignment plus a data-aware SVD initialization), so a residual worry is that opening the matrix helps only in concert with those auxiliaries. A $2\times2$ factorial on the Qwen target (matched $2$B PT tokens, seed 1234) crosses \emph{reachable set} (slice vs.\ full) with \emph{recipe} (LRC vs.\ GPD, a stripped generic recipe: logit-KL only, random-projection init, no activation alignment). Training the full matrix beats the slice under \emph{both} recipes (Table~\ref{tab:crossrecipe}): $+1.82$ under LRC and $+2.86$ under GPD. The gain is a property of \emph{which} parameters are trained, not the surrounding recipe; absolute scores are low by design (one-fifth the tokens; GPD further strips the recipe) and are read only within this matched setting.

\begin{table}[t]
\centering
{\small
\setlength{\tabcolsep}{5pt}
\begin{tabular}{lccc}
\toprule
Recipe & slice & full & gap \\
\midrule
LRC (aux $+$ SVD init) & $58.64$ & $60.46$ & $+1.82$ \\
GPD (KL only $+$ rand) & $39.95$ & $42.81$ & $+2.86$ \\
\bottomrule
\end{tabular}}
\caption{Recipe-independence of the reachable-set gain (Qwen2.5-3B$\to$1.7B, matched $2$B PT, seed 1234). ``slice'' trains only $\mathrm{col}(T)$; ``full'' trains the deployed matrix.}
\label{tab:crossrecipe}
\end{table}

\subsection{The Opened Coordinates Are Functionally Engaged}\label{subsec:engagement}
Decomposing each trained MLP weight against the teacher column space, the Full-MLP LRC student places $6.2\%$ (gate) and $8.3\%$ (up) of its weight energy in $\colperp$ on average---$12$--$16\%$ in the first three layers, decaying with depth---while a plain-LRC student measures $\approx$0.1\%, zero up to numerics (the supplementary material). Surgically removing exactly the complement component ($7.4\%$ of weight energy) collapses the student to chance on all nine tasks; a plain-LRC student confined to $\mathrm{col}(W_T)$ by construction performs normally, and an identity pass through the same pipeline is lossless. The dependence is depth-localized: removing the complement of only the first three layers reproduces most of the collapse ($-31.74$ Avg9); layers $3$--$9$ cost $-2.73$; later layers are within noise. The collapse shows the trained solution is strongly \emph{co-adapted} with the complement component, not that those directions independently encode capability. Separately, RED's representation-rank statistic~\citep{he2026red} dissociates from the gain: the dense and teacher-spectral arms differ $1.7\times$ in early-layer effective rank yet tie on Avg9.

\subsection{Which Comparisons Are Strictly Matched}\label{subsec:scope}
The core reachable-set claims---the $\mathrm{col}(T)$-vs-full gap on each teacher, the equal-parameter and basis controls, the cross-recipe $2\times2$---are strictly matched pairs (same warm start, recipe, data order, budget, seed; only the isolated factor differs). Two comparisons are cross-lineage, both on Qwen: teacher-spectral headline delta ($+10.45$) and implied CORE-vs-Dense margin; both flagged where they appear (\S\ref{sec:limitations}).

\section{Limitations}\label{sec:limitations}

\textbf{(1) No zero-forgetting claim.} Single-layer orthogonality cannot control the end-to-end Jacobian; MMLU does drop.
\textbf{(2) Attribution strength.} The equal-parameter control uses one fixed $A$, one seed; the $2\times2$ runs at $2$B, single-seed (``strongly supports,'' not ``decisively isolates''); the Qwen headline $+10.45$ is cross-lineage, always paired with the strictly matched dense $+6.39$ that the controlled claim rests on. Training runs are single-seed (the field norm; LRC, Minitron, Sheared-LLaMA); lm-eval SE bounds evaluation noise only, not run-to-run variance.
\textbf{(3) A single design axis.} LRC compresses only the hidden dimension; we characterize the optimal $\dmodel$ under an inherited $\dff$, not the optimal $\dff$; the hidden-width sweep lacks per-width no-complement baselines, so its MMLU reading is suggestive.
\textbf{(4) No MoE coverage.} Per-expert expansion ratios sit near or below~$1$: little structural complement.
\textbf{(5) One compression backbone; evaluation scope.} All students use the LRC backbone; external baselines are published numbers, context only---no distillation-SOTA claim; whether other prune--distill methods show the same reserve is untested. Evaluation is $0$-shot multiple-choice/QA (no generative or long-chain reasoning suites). The $\sim$900$\times$ figure is a token count under unmatched recipes, not a compute claim.

\section{Conclusion}\label{sec:conclusion}

Training the full deployed matrix from the identical LRC warm start recovers the stranded capacity at zero added inference cost ($+2.36$/$+2.71$/$+10.45$ Avg9; $2\times$ token efficiency; a half-parameter student matching its teacher), and controls attribute the gain to the reachable set. The audit is not LRC-specific: for any compressed weight, compare the dimension training can reach against the dimension deployment pays for, then justify the gap or close it. Wherever a width is \emph{inherited} rather than deleted, we expect a
reserve of the same kind---testing that on other prune--distill backbones is the natural next step. So: \emph{a weight you deploy but never train is a gap to close---and closing it is free at inference.}

\bibliography{references}

\clearpage
\appendix
\setcounter{table}{0}
\setcounter{figure}{0}
\setcounter{equation}{0}
\setcounter{proposition}{0}
\renewcommand{\thetable}{S\arabic{table}}
\renewcommand{\thefigure}{S\arabic{figure}}
\renewcommand{\theequation}{S\arabic{equation}}
\renewcommand{\theproposition}{S\arabic{proposition}}
\section*{Appendix: Supplementary Material}
\vspace{1.5em}

\section{CORE-LRC: The Teacher-Spectral Realization}\label{app:core}

\emph{Train what you deploy} has a canonical realization, Dense-LRC, and a teacher-spectral realization, \textbf{CORE-LRC}, which reaches the \emph{same} full deployed matrix in the teacher's own SVD basis, with the complement zero-initialized so training starts exactly at the plain-LRC model (main paper, Eq.~2). On the matched Llama settings the two tie; the main text reports whichever realization is stronger per teacher (teacher-spectral on Llama3.2-3B and on the widest Qwen target, where it carries the $+10.45$ headline and the $2\times$ token-efficiency result; canonical dense on Llama3.1-8B). This section restates the construction, gives its mergeability proposition, and details the basis-conditioning analysis.

\begin{figure*}[t]
\centering
\includegraphics[width=\textwidth]{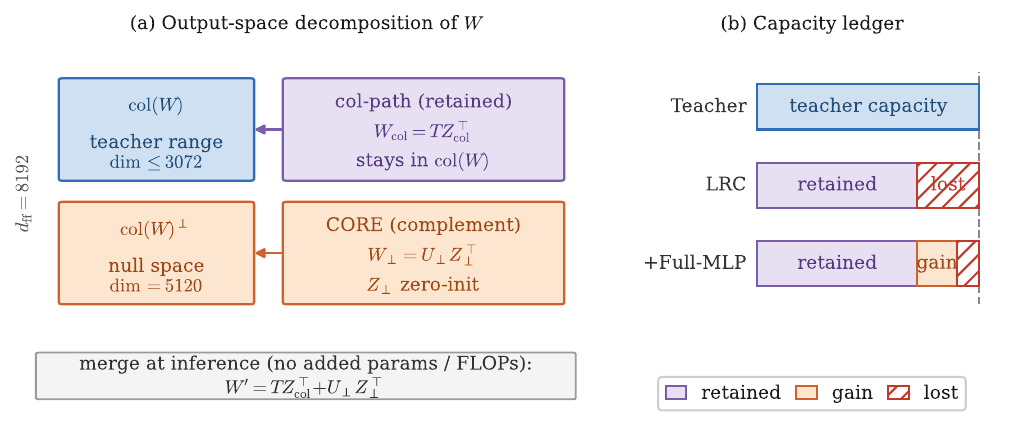}
\caption{CORE-LRC: a complement-coordinate reparameterization. (a) A non-square MLP weight decomposes into the teacher-occupied range and the structural complement; CORE-LRC restricts the update to the complement, zero-initialized and mergeable. (b) Capacity ledger: hidden-dimension compression loses capacity; opening the complement reclaims the unused-but-usable part.}
\label{fig:mech}
\end{figure*}

\paragraph{Construction.}
Let $T\in\mathbb{R}^{\dff\times H}$ be an orthonormal basis of $\mathrm{col}(W)$ and $U_\perp\in\mathbb{R}^{\dff\times(\dff-H)}$ an orthonormal basis of the complement $\colperp$ (both read off the SVD of $W$), so $[\,T\ U_\perp\,]$ is a complete orthonormal basis of $\mathbb{R}^{\dff}$. CORE-LRC augments the LRC column path with a complement path,
\begin{equation}\label{eq:ocd-supp}
W_{\mathrm{student}} = T\,Z_{\mathrm{col}}^{\top} + U_\perp\,Z_\perp^{\top},
\end{equation}
with $Z_{\mathrm{col}}\in\mathbb{R}^{r\times H}$ and $Z_\perp\in\mathbb{R}^{r\times(\dff-H)}$; $Z_\perp$ is zero-initialized, so the complement path vanishes at the start and the student is exactly the plain-LRC model. For the down projection the dual construction fills the right complement. At inference the weight merges to a single $\dff\times r$ matrix of the plain-LRC shape: no added parameters or FLOPs (Figure~\ref{fig:mech}).

\begin{proposition}[Mergeability]\label{prop:moca}
Let $U_\perp^{\top}T=\mathbf{0}$ with $[\,T\ U_\perp\,]$ orthonormal. For the merged weight of Eq.~\eqref{eq:ocd-supp}:
\emph{(i)} the two paths write to orthogonal output subspaces, and at $Z_\perp{=}\mathbf{0}$ the perp path vanishes, so the student is unchanged at initialization;
\emph{(ii)} $W_{\mathrm{student}}$ keeps the plain-LRC shape, so after merging the deployed model adds no parameters or FLOPs;
\emph{(iii)} the two path gradients read $G$ in complementary coordinates ($G^{\top}T$ and $G^{\top}U_\perp$), so updating one path cannot change the other's output.
\end{proposition}
\noindent All three claims follow from orthonormality; the proposition constrains only the single-layer MLP output (end-to-end behavior through later layers is not controlled; Limitation~1). Because $[\,T\ U_\perp\,]$ is complete, the two paths span the entire deployed weight space: with $Q{=}[\,T\ U_\perp\,]$ and $M{=}[Z_{\mathrm{col}}^{\top};Z_\perp^{\top}]$, CORE-LRC is $W{=}QM$---the standard student weight trained in a fixed teacher-aligned basis, functionally equivalent to Dense-LRC and differing only in the coordinate system AdamW optimizes in.

\paragraph{Basis conditioning: when the basis starts to matter.}
On the matched narrow-teacher settings Dense-LRC and CORE-LRC tie ($65.27$ vs.\ $65.40$ Avg9)---basis-independence. The wide-MLP Qwen2.5-3B target, whose residual stream carries massive-activation channels~\citep{sun2024massive}, is where the two part company: the teacher-spectral basis realizes far more of the same expansion at a fixed $10$B budget---Avg9 $63.44$ for CORE-LRC ($+10.45$ over the matched plain-LRC baseline) versus $59.38$ for the canonical-dense arm ($+6.39$), with CORE-LRC matching the original recipe's $\sim$20B-token baseline ($63.43$) at half the tokens ($2\times$ token efficiency). Because the CORE-Qwen arm comes from a different training run than the seed-matched Dense/LRC pair, we read the cross-realization margin ($\approx$$+4$) \emph{qualitatively}---consistent with AdamW's coordinate dependence~\citep{zhang2025adamrotation}---while the same-lineage dense delta ($+6.39$) carries the strictly controlled reachable-set claim; a matched pair on a wide-MLP target is left to future work. The token-count observation that CORE-LRC at $10$B reaches plain LRC's original-recipe $\sim$20B accuracy ($63.44$ vs.\ $63.43$) is a token-count comparison, \emph{not} a controlled compute-to-target estimate, and we do not report one.

\section{Additional Result Tables}\label{app:tables}

\begin{table*}[t]
\centering
{\small
\setlength{\tabcolsep}{3.2pt}
\begin{tabular}{lccccccccccc}
\toprule
Model & ARC-E & ARC-C & LogiQA & CSQA & PIQA & WinoG & BoolQ & SciQ & MMLU & \textbf{Avg9} & MathQA \\
\midrule
Teacher Llama3.1-8B & 82.28 & 55.55 & 31.64 & 76.17 & 80.20 & 73.56 & 85.44 & 97.30 & 68.31 & \textbf{72.27} & 39.53 \\
\textit{Ref.\ Llama3.2-3B-Instruct} ($\sim$9T tok) & 73.86 & 46.25 & 29.65 & 67.97 & 75.56 & 68.03 & 78.53 & 95.30 & 60.56 & \textit{66.18} & 34.84 \\
\midrule
LRC-2.7B 10B PT       & 77.48 & 47.61 & 33.79 & 66.91 & 72.58 & 63.22 & 78.23 & 96.10 & 54.00 & 65.55 & 26.53 \\
\midrule
Dense-LRC 2.7B 10B PT        & 80.89 & 51.37 & 31.64 & 74.12 & 74.21 & 67.40 & 79.36 & 96.80 & 58.55 & \textbf{68.26} & 29.78 \\
Dense-LRC 2.7B 10B PT$+$SFT  & 80.18 & 49.83 & 31.49 & 78.79 & 75.79 & 66.54 & 82.57 & 96.60 & 58.54 & \textbf{68.93} & 32.43 \\
\midrule
CORE-LRC$^{\ddagger}$ 2.7B 10B PT        & 79.59 & 50.09 & 31.49 & 74.20 & 73.01 & 64.64 & 79.57 & 95.90 & 58.65 & 67.46 & 29.82 \\
CORE-LRC$^{\ddagger}$ 2.7B 10B PT$+$SFT  & 78.20 & 50.77 & 31.64 & 78.54 & 73.61 & 67.25 & 83.61 & 95.70 & 59.70 & 68.78 & 32.73 \\
CORE-LRC$^{\ddagger}$ 2.7B 16.5B PT$+$SFT& 78.32 & 50.34 & 32.10 & 78.87 & 74.65 & 67.17 & 83.94 & 95.40 & 60.20 & \textbf{69.00} & 33.60 \\
\midrule
$\Delta$ (Dense-LRC$-$LRC, 10B PT) & +3.41 & +3.76 & $-$2.15 & +7.21 & +1.63 & +4.18 & +1.13 & +0.70 & +4.55 & \textbf{+2.71} & +3.25 \\
\bottomrule
\end{tabular}}
\caption{Llama3.1-8B$\to$2.7B: training the deployed matrix vs.\ a matched LRC baseline and the teacher. The matched full-matrix arm (Dense-LRC realization) adds $+2.71$ Avg9 / $+4.55$ MMLU at 10B PT. The \emph{Ref.} row is the official Llama3.2-3B (Meta's own $\sim$9T-token prune--distill compression of the \emph{same} teacher), which our $2.7$B student exceeds on Avg9 at $\sim$900$\times$ less PT. $^{\ddagger}$Teacher-spectral realization from a separate training lineage (Appendix~\ref{app:core}); context, not a matched pair. Avg9 excludes MathQA.}
\label{tab:llama8}
\end{table*}

\begin{table*}[t]
\centering
{\small
\setlength{\tabcolsep}{3.2pt}
\begin{tabular}{lccccccccccc}
\toprule
Model & ARC-E & ARC-C & LogiQA & CSQA & PIQA & WinoG & BoolQ & SciQ & MMLU & \textbf{Avg9} & MathQA \\
\midrule
Teacher Qwen2.5-3B & 77.06 & 47.95 & 31.49 & 78.71 & 77.37 & 69.22 & 80.15 & 94.70 & 65.38 & \textbf{69.11} & 35.51 \\
\midrule
LRC-1.7B 10B PT (matched)      & 64.48 & 38.48 & 28.88 & 38.66 & 67.52 & 50.91 & 67.19 & 84.40 & 36.42 & 52.99 & 25.46 \\
LRC-1.7B $\sim$20B PT (orig.\ budget) & 69.49 & 42.75 & 33.26 & 70.27 & 71.38 & 63.85 & 75.78 & 89.00 & 55.13 & 63.43 & -- \\
\midrule
Dense-LRC 1.7B 10B PT       & 72.90 & 42.32 & 27.96 & 60.28 & 69.48 & 55.09 & 69.63 & 86.60 & 50.13 & \textbf{59.38} & 27.71 \\
Dense-LRC 1.7B 10B PT$+$SFT & 71.55 & 40.87 & 28.11 & 58.72 & 68.99 & 55.33 & 74.31 & 84.40 & 46.43 & 58.75 & 29.15 \\
\midrule
CORE-LRC$^{\ddagger}$ 1.7B 10B PT        & 74.79 & 44.45 & 29.95 & 69.78 & 72.31 & 60.54 & 75.35 & 90.38 & 53.45 & \textbf{63.44} & 32.76 \\
CORE-LRC$^{\ddagger}$ 1.7B 10B PT$+$SFT  & 75.04 & 44.97 & 29.03 & 69.86 & 73.29 & 60.62 & 76.15 & 91.70 & 53.68 & \textbf{63.82} & 33.23 \\
\midrule
$\Delta$ (Dense$-$LRC, 10B PT, matched)  & +8.42 & +3.84 & $-$0.92 & +21.62 & +1.96 & +4.18 & +2.44 & +2.20 & +13.71 & \textbf{+6.39} & +2.25 \\
$\Delta$ (CORE$-$LRC, 10B PT)$^{\ddagger}$ & +10.31 & +5.97 & +1.07 & +31.12 & +4.79 & +9.63 & +8.16 & +5.98 & +17.03 & \textbf{+10.45} & +7.30 \\
\bottomrule
\end{tabular}}
\caption{Qwen2.5-3B$\to$1.7B: the widest MLP is where training what you deploy pays off most. Teacher-spectral CORE-LRC reaches $63.44$ at $10$B PT ($+10.45$ over the matched baseline)---matching the original $\sim$20B-token baseline ($63.43$) at half the tokens ($2\times$ token efficiency), $63.82$ after SFT. Canonical Dense-LRC recovers $+6.39$ in a strictly matched same-lineage comparison---the fully controlled attribution evidence. On this widest, most ill-conditioned target the teacher-spectral basis is the markedly better optimizer (CORE $\approx$4 above Dense), in contrast to the Llama targets where the two tie. $^{\ddagger}$The CORE-Qwen arm is a separate training lineage from the seed-matched Dense/LRC pair, so its $\Delta$ row is read jointly with the matched Dense row. The short general SFT does \emph{not} help the under-converged Dense arm ($58.75$ vs.\ PT $59.38$); we read the Qwen result at PT. Avg9 excludes MathQA; ``--'' = not evaluated.}
\label{tab:qwen}
\end{table*}

\begin{table}[t]
\centering
{\small
\setlength{\tabcolsep}{4pt}
\begin{tabular}{lcccc}
\toprule
Model & Params & PT tok & \textbf{Avg9} & MMLU \\
\midrule
Sheared-LLaMA & 2.7B & 50B  & 52.55 & 26.56 \\
Minitron      & 4B   & 94B  & 68.31 & 56.77 \\
\midrule
MiniCPM-1.2B   & 1.2B & 1T  & 60.42 & 44.23 \\
InternLM2-1.8B & 1.8B & 2T  & 62.60 & 43.75 \\
SmolLM2-1.7B   & 1.7B & 11T & 60.50 & 48.50 \\
Qwen3-1.7B     & 1.7B & 36T & 63.17 & 55.44 \\
\midrule
LRC-1.5B \emph{(Llama3.2-3B)} & 1.5B & 10B & 63.48 & 49.42 \\
LRC-1.7B \emph{(Qwen2.5-3B)} & 1.7B & 20B & 64.98 & 54.93 \\
\textbf{Full-MLP LRC 1.5B} & 1.5B & 10B & \textbf{66.21} & 54.86 \\
\textbf{Full-MLP LRC 2.7B} & 2.7B & 10B & \textbf{68.93} & 58.54 \\
\bottomrule
\end{tabular}}
\caption{External context (published values reported by LRC under the same $0$-shot harness and nine tasks; teachers, budgets, and sizes differ---context, not a controlled comparison). Our two rows report the stronger realization per setting: teacher-spectral for the $1.5$B student ($66.21$; Dense-LRC $65.95$), canonical dense for the $2.7$B student ($68.93$; teacher-spectral $69.00$ at $16.5$B). Our rows are $10$B-PT$+$SFT students; the PT-tok column counts the PT budget only (the short SFT adds $\approx$0.62B tokens).}
\label{tab:external}
\end{table}

\begin{table}[t]
\centering
{\small
\setlength{\tabcolsep}{2.6pt}

\begin{tabular}{lccccccc}
\toprule
PT tokens & 1B & 2B & 3B & 5B & 8B & 10B & 16.5B \\
\midrule
Avg9 & 60.02 & 62.75 & 63.20 & 64.11 & 64.53 & 65.40 & 65.35 \\
\bottomrule
\end{tabular}}
\caption{PT-token saturation (Llama3.2-3B$\to$1.5B; one epoch; teacher-spectral realization). The sweep flattens at about $10$B; $16.5$B is within noise of $10$B.}
\label{tab:sat}
\end{table}

\paragraph{Recovering capacity along the hidden axis.}
Holding the teacher (Llama3.2-3B), data, budget ($10$B, one epoch), and losses fixed, we sweep the student hidden dimension $\dmodel^{(S)}$ from $1024$ to $3072$ with the MLP fully unfrozen (teacher-spectral realization; Table~\ref{tab:hidden}). The 9-task average rises monotonically and crosses the teacher near $\dmodel^{(S)}\!\approx\!2048$ (a standard distillation effect~\citep{furlanello2018born}); MMLU, stuck at $54.67$ at $\dmodel^{(S)}{=}1536$ despite full-matrix training, climbs monotonically and reaches the teacher's $60.56$ only near the native width. This \emph{associates} the residual MMLU gap with overall student width: with the full MLP complement already open, more MLP-tail capacity does not close it. Widening $\dmodel^{(S)}$ moves several capacity axes at once, so we read this as suggestive rather than a clean isolation; matched no-complement baselines at each width are future work.

\begin{table}[t]
\centering
{\small
\setlength{\tabcolsep}{4pt}
\begin{tabular}{cccccc}
\toprule
$\dmodel^{(S)}$ & $\rhoeff$ & Avg9 & MMLU & $\Delta$Avg9 & $\Delta$MMLU \\
\midrule
1024 & 8.00 & 62.01 & 48.87 & $-4.17$ & $-11.69$ \\
1536 & 5.33 & 65.40 & 54.67 & $-0.78$ & $-5.89$ \\
2048 & 4.00 & 66.43 & 56.14 & $+0.25$ & $-4.42$ \\
2560 & 3.20 & 67.76 & 58.87 & $+1.58$ & $-1.69$ \\
3072 & 2.67 & 68.02 & 62.16 & $+1.84$ & $+1.60$ \\
\midrule
\textit{teacher} & 2.67 & 66.18 & 60.56 & --- & --- \\
\bottomrule
\end{tabular}}
\caption{Full-MLP LRC across student hidden sizes (Llama3.2-3B teacher; $10$B PT, one epoch; teacher-spectral realization). MMLU recovers toward the teacher only as $\dmodel^{(S)}$ grows.}
\label{tab:hidden}
\end{table}

\begin{figure*}[t]
\centering
\includegraphics[width=\textwidth]{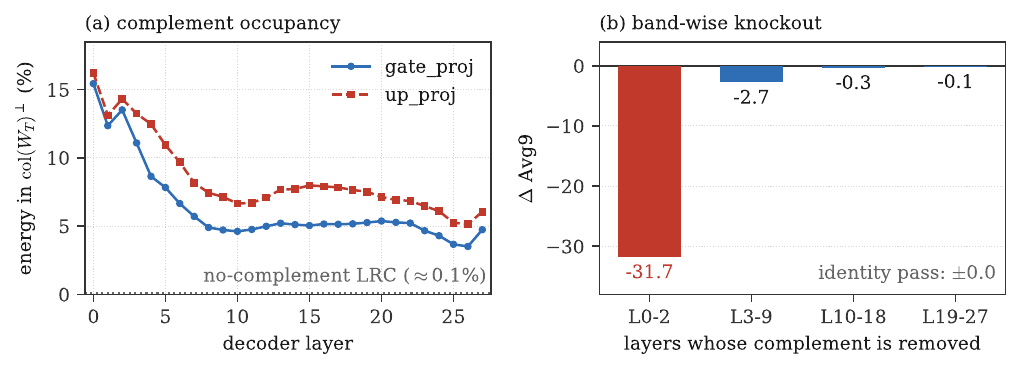}
\caption{The opened coordinates are functionally engaged. (a)~Per-layer fraction of MLP weight energy in the teacher-orthogonal complement (Llama3.2-3B$\to$1.5B student): highest in the first three layers ($12$--$16\%$), decaying with depth; a plain-LRC student measures $\approx$0.1\% (dotted). (b)~Surgically removing the complement component of only the layers in each band: the first three layers carry nearly the entire functional dependence ($-31.74$ Avg9, collapse to chance); an identity pass is lossless.}
\label{fig:engagement}
\end{figure*}

\section{Implementation Details}\label{app:impl}

\paragraph{Student construction and initialization.}
The student is initialized by a greedy, calibration-data-aware SVD pass over a few hundred packed $2048$-token sequences: an SVD of the (tied) embedding gives the master projection $P\in\mathbb{R}^{r\times H}$ compressing the residual stream; the per-layer zoom matrices are initialized by ridge regression against teacher activations, propagating the student's own compressed hidden states forward so each layer is fit on the distribution it will actually see. The Llama students use a plain SVD ($k{=}0$). Qwen2.5 carries a handful of massive-activation residual channels ($\sim$$10^3$)~\citep{sun2024massive}; for the Qwen student an \emph{outlier-aware} $P$ optionally preserves the top-$k$ ($k{=}48$) outlier channels verbatim as identity rows and SVD-compresses the rest---an initialization-side conditioning option, orthogonal to the reachable-set intervention, worth $+0.68$ Avg9 (main paper). All other LRC system switches (KV-Direct, embedding/LM-head recovery, Wiener initialization) are disabled in every reported run.

\paragraph{Objective.}
The per-token loss is
\begin{equation*}
\begin{aligned}
\mathcal{L}={}& w_{\mathrm{KL}}T^{2}\,\mathrm{KL}\!\big(\sigma(z_{\mathcal{T}}/T)\,\|\,\sigma(z_{\mathcal{S}}/T)\big)\\
&{}+w_{\mathrm{NTP}}\,\mathrm{CE}(y,z_{\mathcal{S}})+\lambda_{\mathrm{aux}}\mathcal{L}_{\mathrm{align}},
\end{aligned}
\end{equation*}
where $\mathcal{L}_{\mathrm{align}}$ averages per-layer hidden-state MSE (mapped to teacher width via $P$), per-head attention cosine alignment on $Q/K/V/O$, and MLP-intermediate MSE on the post-SwiGLU product. All reported PT runs use $w_{\mathrm{KL}}{=}w_{\mathrm{NTP}}{=}1.0$, $T{=}40$, $\lambda_{\mathrm{aux}}{=}0.2$; these are held identical for plain LRC, Dense-LRC, and CORE-LRC, so they do not differentiate any reported comparison.

\paragraph{Training and evaluation.}
AdamW (cosine schedule, $10\%$ warmup, grad-clip $1.0$), learning rate $1{\times}10^{-4}$ (PT) and $1{\times}10^{-5}$ (SFT), effective batch $32$ sequences of length $2048$, bf16, one epoch per budget. PT data: $\approx$10B FineWeb-Edu tokens (educational score $\geq$4) plus $\approx$0.35B OpenHermes tokens, tokenized with each teacher's own tokenizer. SFT: $\approx$0.62B tokens over an 8-dataset general mixture. Evaluation: standard $0$-shot accuracy via the LM Evaluation Harness for all models; Avg9 is the mean over ARC-Easy, ARC-Challenge, LogiQA, CommonsenseQA, PIQA, WinoGrande, BoolQ, SciQ, and MMLU.

\paragraph{Training-time cost.}
Relative to plain LRC, both realizations roughly double the trainable parameter count ($\approx$0.40B$\to$$\approx$1.06B for the 3B setting); the per-token MLP FLOPs are identical for plain LRC and Dense-LRC (same deployed shape), and the teacher-spectral realization adds a fixed-basis reconstruction measured at $\approx$12\% step time over Dense-LRC on matched hardware, plus $7$--$28$ GB of frozen bases in bf16. All of this is training-only: after merging, the deployed student has exactly the plain-LRC parameter count and FLOPs. We do not report a controlled compute-to-target estimate (Appendix~\ref{app:core}).

\end{document}